\documentclass[final,5p,times,twocolumn,authoryear]{elsarticle}

\usepackage{amssymb}
\usepackage{amsmath}
\usepackage{multirow}
\usepackage{xcolor}
\usepackage{hyperref}
\hypersetup{
  colorlinks=true, 
  linkcolor=cyan, 
  citecolor=cyan, 
  filecolor=cyan, 
  urlcolor=cyan, 
  pdfborder={0 0 0},
}
\usepackage{comment}
\usepackage{lineno}
\usepackage{algorithm}
\usepackage{algpseudocode}

\begin{document}

\begin{frontmatter}

\title{Supervised domain adaptation for building extraction from off-nadir aerial images}

\author[inst1,inst2]{Bipul Neupane\corref{cor1}}
\ead{bneupane@student.unimelb.edu.au}
\author[inst1,inst2]{Jagannath Aryal}
\ead{jagannath.aryal@unimelb.edu.au}
\author[inst2]{Abbas Rajabifard}
\ead{abbas.r@unimelb.edu.au}

\cortext[cor1]{Corresponding author}
\affiliation[inst1]{organization={Earth Observation and AI Research Group, Faculty of Engineering and IT, The University of Melbourne},
            addressline={Parkville}, 
            city={Melbourne}, 
            postcode={3010}, 
            state={VIC},
            country={Australia}}

\affiliation[inst2]{organization={Department of Infrastructure Engineering, Faculty of Engineering and IT, The University of Melbourne},
            addressline={Parkville}, 
            city={Melbourne},
            postcode={3010}, 
            state={VIC},
            country={Australia}}

\begin{abstract}
Building extraction $-$ needed for inventory management and planning of urban environment $-$ is affected by the misalignment between labels and off-nadir source imagery in training data. Teacher-Student learning of noise-tolerant convolutional neural networks (CNNs) is the existing solution, but the Student networks typically have lower accuracy and cannot surpass the Teacher's performance. This paper proposes a supervised domain adaptation (SDA) of encoder-decoder networks (EDNs) between noisy and clean datasets to tackle the problem. EDNs are configured with high-performing lightweight encoders such as EfficientNet, ResNeSt, and MobileViT. The proposed method is compared against the existing Teacher-Student learning methods like knowledge distillation (KD) and deep mutual learning (DML) with three newly developed datasets. The methods are evaluated for different urban buildings (low-rise, mid-rise, high-rise, and skyscrapers), where misalignment increases with the increase in building height and spatial resolution. For a robust experimental design, 43 lightweight CNNs, five optimisers, nine loss functions, and seven EDNs are benchmarked to obtain the best-performing EDN for SDA. The SDA of the best-performing EDN from our study significantly outperformed KD and DML with up to 0.943, 0.868, 0.912, and 0.697 F1 scores in the low-rise, mid-rise, high-rise, and skyscrapers respectively. The proposed method and the experimental findings will be beneficial in training robust CNNs for building extraction.
\end{abstract}


\end{frontmatter}

\section{Introduction}
\begin{figure*}[!ht]
  \centering
  \includegraphics[width=0.8\linewidth]{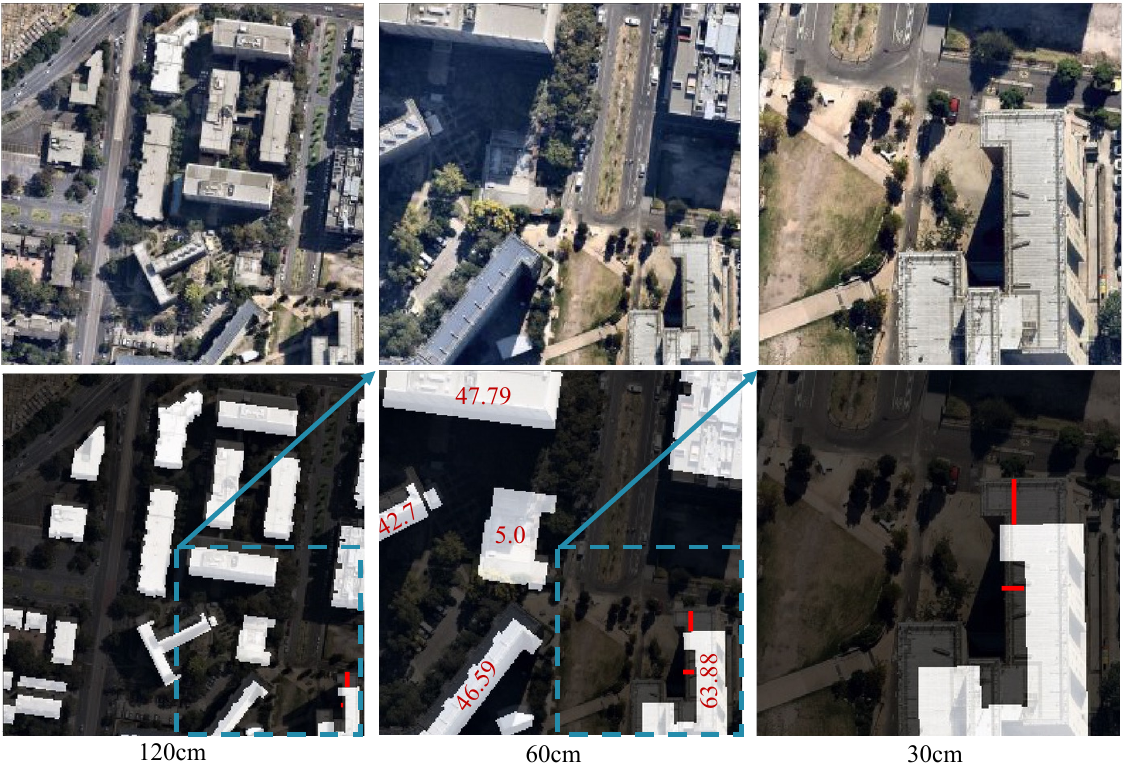}
  \caption{Misalignment of building labels with an increase in spatial resolution of aerial image and height of the building. The numbers in the labels show the height of the building and the length of the red line shows the displacement of labels. As an illustration, the building with a height of 63.88 m in 120 cm (left) spatial resolution image is displaced less in comparison with 60 cm (middle) and 30 cm (right) spatial resolution images. The images are collected from \textit{Nearmap Tile API} over the City of Melbourne, Australia.}
      \label{fig:misalignment}
      \end{figure*}

Building extraction from high-resolution remote sensing (RS) images provides essential data for urban building inventories \citep{li2022review}. These inventories enhance the management of urban environment and improve societal aspects of urban living with applications such as mapping, urban planning, disaster management, sprawl and green space management, urban heat monitoring, change detection, and humanitarian efforts \citep{schunder2020spatial, herfort2021evolution}. With the advancement in object detection and semantic segmentation \citep{neupane2024cnns}, urban buildings are extracted from RS images with EDN networks \citep{amirgan2024semantic}. However, the extraction poses a significant challenge due to misalignment caused by variations in building heights and off-nadir angles of airborne sensors. The misalignment increases along with the increase in spatial resolution and building height as depicted in Figure \ref{fig:misalignment}. This reduces the accuracy of high-rise urban building extraction methods. This issue is under investigation (see SpaceNet 4 Challenge \citep{weir2019spacenet}), as global data providers like Microsoft's Building footprints \citep{huang2021100}, Google's Open buildings \citep{sirko2021continental}, ESRI\footnote{See: \href{https://www.arcgis.com/home/item.html?id=4e38dec1577b4b7da5365294d8a66534}{ESRI's Deep learning model to extract building footprints}} and local industries like Geoscape\footnote{See page 23 of \href{https://geoscape.com.au/wp-content/uploads/2022/03/Buildings-Product-Guide-v3.1.pdf}{Geoscape's Building Product Guide Version 3.1.}} have reported that this issue has impacted their data products.

Few methods have been devised to address the problem of misalignment. Relief displacement correction \citep{chen2021extracting}, multi-task learning \citep{li2022poi}, and learning from depth \citep{zhu2021depth} are studied as potential solutions. The segmented outputs in these methods are improved with several post-processing steps including morphological operations that limit their robustness and end-to-end usage. Apart from tackling the misalignment problem, some studies distinguish between facade and roof in a multi-view oblique image \citep{zhu2021depth, yan2023building, hu2024building}. The problem of extracting roofs from multi-view oblique images is not the focus of this work. \citep{wang2022bonai} developed a building dataset with labels for roof, footprints, and offset between them and employed a learning offset vector (LOFT) method with CNN that learned from the offset. Despite improvements, the accuracy of the LOFT remains to be improved. Besides these methods, the other studies investigate the knowledge-transfer methods to improve CNN's tolerance to misalignment. Such a method features an initial pre-training of CNNs on a large misaligned dataset, followed by distillation of its knowledge into a smaller CNN trained with a smaller error-free clean dataset. \citep{xu2022improving} used KD to train a Teacher network on the large noisy data (with misalignment) and distil a Student using a limited set of clean data (without misalignment). Here, the Teacher served as a guide, enabling the Student to tap into its knowledge for noise-tolerant learning. They compare the performance of KD against DML \citep{zhang2018deep}, which allows an ensemble of Students to learn collaboratively and teach each other throughout the distillation from the Teacher. \citep{neupane2023knowledge} extended the investigation of KD with noisy data of off-nadir images and clean data of ortho-rectified images. From these studies, the Teacher-Student learning of EDNs improved the Student's tolerance to misalignment.

Among the solutions, training noise-tolerant networks provides a promising direction for addressing the misalignment problem. However, these methods such as KD were originally designed to compress the knowledge of large-immovable Teachers into the small-deployable Students trained for a specific task \citep{hinton2015distilling}. Students are smaller deployable networks that cannot store as much knowledge or surpass the performance of the Teacher network. With this limitation, Students lack the accuracy and precision in segmenting the complex urban buildings.

This paper proposes SDA to train misalignment-tolerant models. SDA is originally a transfer learning method that allows adapting a model trained on a source domain to perform efficiently on a target domain, using labelled data from both domains. \citep{aryal2023multi} used SDA to adapt an EDN that is pre-trained on a building dataset from Christchurch, New Zealand, to a new dataset with complex urban buildings. \citep{alexander2024pre} fine-tuned regional models for performance gains in building roof extraction. Unlike KD and DML, SDA allows the model to be re-trained to outperform the pre-trained version. SDA has yet to be explored for addressing the misalignment problem. We adopt the method of SDA to pre-train an EDN using a larger Teacher dataset with misalignment (source domain) and subsequently adapt the model on a smaller, error-free Student dataset (target domain). 

The main contributions of this study are:

\begin{itemize}
    \item SDA is proposed to tackle the misalignment between the building's image-label pairs due to the off-nadir angle of source images. It is compared against KD and DML for four building types (low-rise, mid-rise, high-rise, and skyscrapers) and three spatial resolutions of images.
    
    \item Three new urban building datasets (with high-risers and skyscrapers) are developed with ``large-noisy'' and ``small-clean'' image-label pairs to train and evaluate the method. The codes to access and reproduce the datasets are available at: \href{https://github.com/bipulneupane/TeacherStudentBuildingExtractor}{https://github.com/bipulneupane/TeacherStudentBuilding\\Extractor}.
    
    \item A benchmark of 43 lightweight CNNs, 5 optimisers, 9 loss functions, and 7 EDNs is presented to identify the best-performing configuration for SDA and to compare against KD and DML. The CNNs include lightweight CNNs from Google, Apple, Meta, Amazon, and Huawei. To the best of our knowledge, such extensive benchmarks of lightweight CNNs and some of their integration into an EDN are novel for building extraction.
\end{itemize}

\section{Materials and methods}\label{sec:method}

\subsection{Data Preparation}\label{sec:meth:dataset}
Four datasets are used in this study: one benchmark dataset for finding the best hyperparameters, CNNs, and EDNs, and three new datasets developed for training, adapting/distilling, and evaluating EDNs. The benchmark dataset identifies optimal configurations to set up the Teacher (pre-trained network) and Students (adapted/distilled networks) for SDA. Afterwards, the Teacher is trained and evaluated on the Teacher's dataset (T), while the Students are adapted/distilled and evaluated using the Student's dataset (S) and Evaluation dataset (Ev). The Students' ability to minimise misalignment is determined by their scores on these two datasets.

\begin{itemize}
    \item \textit{Benchmark dataset:} A subset of the Massachusetts Building dataset provided by \citep{neupane2024cnns} is used for the preliminary investigation. It consists of image tiles of 256x256 pixels and the train-validation image ratio is 800-160.
    
    \item \textit{Teacher's dataset (T)} is developed by masking and tiling the building polygons provided by the City of Melbourne\footnote{See: \href{https://data.melbourne.vic.gov.au/explore/dataset/2020-building-footprints/information/}{2020 Building Footprints from the City of Melbourne}}. Multi-resolution image tiles (256x256 pixels) of 30, 60, and 120 cm spatial resolution are collected using \textit{Nearmap Tile API}. The misalignment between these image-label pairs is not accounted for, making it the ``large-noisy'' Teacher's dataset. The ratio of train-validation images is 6626-643.
    
    \item \textit{Student's dataset (S)} is developed for adaptation/distillation purposes. Multi-resolution image tiles and an ortho-rectified mosaic are collected from \textit{Nearmap} for the central business district (CBD) of Melbourne. The labels are manually annotated with a reference to the building roof samples provided by the City Council. The ratio of train-validation images is 746-243.
    
    \item \textit{Evaluation dataset (Ev)} is curated as a dataset to evaluate the adapted/distilled networks on the manually prepared ground truths (GTs) for image scenes of four building types: low-rise, mid-rise, high-rise, and skyscrapers. This dataset does not have training samples (only validation). A subset of 55 image tiles of T is taken and manually annotated. The building types are separated according to their heights as defined by the Australian Bureau of Statistics (ABS)\footnote{See: \href{https://www.abs.gov.au/ausstats/abs@.nsf/Lookup/8752.0Feature+Article1Dec\%202018}{Characteristics of Apartment Building Heights by ABS}}.
\end{itemize}

\begin{figure*}[!ht]
  \centering
  \includegraphics[width=\linewidth]{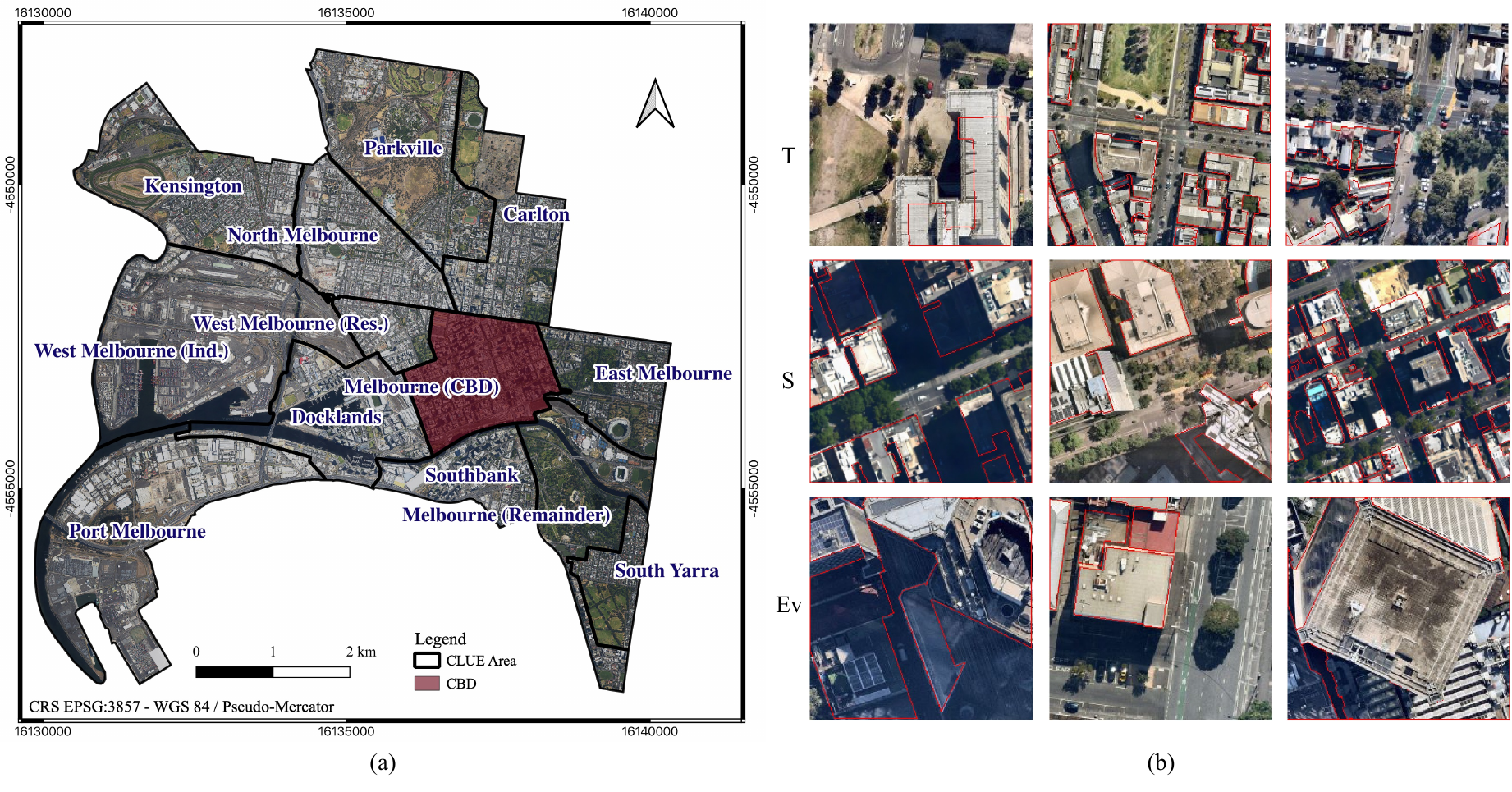}
  \caption{Study area and sample image-label pairs of the Teacher's dataset (T), Student's dataset (S), and Evaluation dataset (Ev). (a) City of Melbourne's Census of Land Use and Employment (CLUE) boundaries and CBD area. (b) Sample image-label pairs.}
      \label{fig:samples}
      \end{figure*}

In summary, (i) T contains misaligned images and labels on both training and validation samples, (ii) S contains off-nadir images with manually annotated labels and ortho-rectified images with aligned labels of complex high-rise buildings, (iii) Ev contains off-nadir images of high-rise buildings from T and manually annotated samples. The study area and the samples of the T, S, and Ev datasets are shown in Figure \ref{fig:samples}.

\subsection{Supervised domain adaptation}\label{sec:meth:sda}
\begin{figure*}[!ht]
  \centering
  \includegraphics[width=\linewidth]{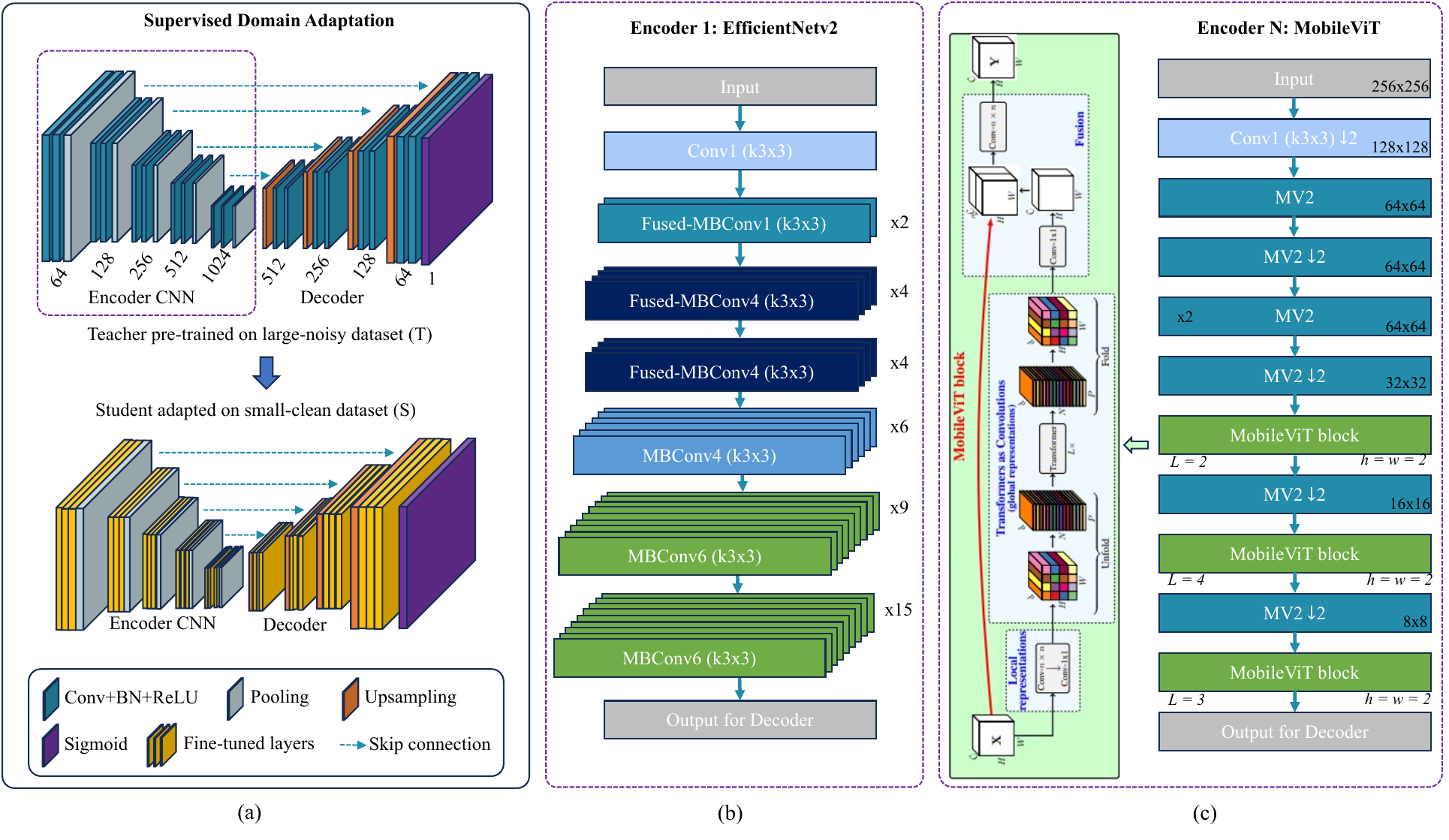}
  \caption{An illustrative example of the SDA with EDN formulated with different options of encoder CNN. (a) SDA of a U-Net EDN with all layers of the pre-trained Teacher being adapted to the Student dataset. (b) Encoder option 1 of EfficientNetv2 \citep{tan2021efficientnetv2} CNN from Google Brain. (c) Encoder option N of MobileViT \citep{mehta2021mobilevit} CNN from Apple.}
      \label{fig:sda}
      \end{figure*}

SDA is a transfer learning method where samples are available in both the source and the target domain \citep{saha2011active}. The idea is to adapt (fine-tune) a DL model pre-trained on a source dataset to a different target dataset by executing one more level of training on the new dataset. We adopt the SDA to pre-train an EDN on the T dataset with misalignment (source domain) and fine-tune the model on the S dataset with clean labels (target domain). All layers of the pre-trained model (Teacher) are fine-tuned to keep the learning from T while updating them based on S. Fine-tuning all layers allows the model to retain the beneficial general features learned during pre-training in T while adapting to the specific nuances of S. An illustration of a U-Net EDN under our SDA setting is shown in Figure \ref{fig:sda}(a).

\subsection{Integration of CNNs to prepare the EDNs for SDA} \label{sec:meth:architecture}
EDNs allow the plug-and-play use of backbone CNNs (encoders) to extract multi-scale feature information from input images. U-Net commonly used EDN for semantic segmentation of buildings \citep{neupane2024cnns}. It processes an input image ($256 \times 256 \times 3$) through the encoder CNNs, while the decoder part upsamples the features learned from the encoder up to the final segmented output. Skip connections transport feature maps from the encoder to corresponding decoder blocks for concatenation. Figure \ref{fig:sda}(a) illustrates the U-Net architecture with an example of two encoder CNNs: EfficientNetv2 (Figure \ref{fig:sda}(b)) and MobileViT (Figure \ref{fig:sda}(c)). To use these CNNs as the encoder of U-Net, we remove their final layers of 1x1 convolution and fully connected layers. The remaining layers are then connected to the decoder that comprises 2D Convolution, batch normalisation (BN), ReLU, and attention (we use this optional layer) layers.

\textit{EfficientNetv2} \citep{tan2021efficientnetv2} is an improved lightweight CNN from Google. It uses Fused-MBConv convolutional blocks alongside the NAS component, and like its predecessor v1, it offers multiple scaled versions. The network uses different variations of Mobile Inverted Bottleneck Convolution (MBConv) blocks: MBConv1, MBConv4, and MBConv6. These blocks are a type of inverted residual block central to the network architecture, providing efficient and flexible building blocks for CNNs. The blocks consist of the EdgeResidual layer to improve the flow of gradients during back-propagation, the non-linear activation function SiLU to introduce non-linearity, and the Squeeze-and-Excitation layer to improve the representational power of a network by explicitly modelling the inter-dependencies between the channels of convolutional features.

\textit{MobileViT} \citep{mehta2021mobilevit} is a compact vision transformer developed for mobile devices. These models use MobileNetv2 (MV2) blocks to leverage their efficiency and lightweight nature for mobile and edge devices. MV2 blocks consist of an expansion layer, depthwise convolution, ReLU6 activation for non-linearity, pointwise convolution to project features back to the desired output dimension, and residual connection between input and output. It uses an additional MobileViT block which is specialised to integrate the strengths of CNNs and transformers. It consists of the convolutional layer that extracts local features, the Transformer layer to capture global context and long-range dependencies, and the Reconstruction layer that reshapes and passes the output of the Transformer layer through another convolutional layer (to combine the local and global information). MobileViT is available in several smaller versions (s, xs, and xxs).

Other CNNs are described in our GitHub\footnote{See: \href{https://github.com/bipulneupane/TeacherStudentBuildingExtractor}{Description of other CNNs.}}, along with the codes to regenerate the datasets developed in this study.

\subsection{Investigating hyperparameters, CNNs, and EDNs for SDA}\label{sec:meth:selection}
SDA can be performed with any configuration of EDN and CNNs. As we aim to study SDA and compare the results to model compression methods of KD and DML (described in supplementary material), we look for a large Teacher network and multiple lightweight Student networks (with $<$20 million network parameters). U-Net with VGG19 CNN (U-VGG19 in short) has higher network parameters and is the high-scorer in the recent-most benchmarks \citep{neupane2024cnns}. U-VGG19 can be used as a Teacher network. To the best of our knowledge, there are no extensive benchmarks of lightweight CNNs that can configure the Students. Therefore, to identify the best-performing lightweight EDN configurations for the Students, a preliminary investigation of hyperparameters, lightweight CNNs, and EDN networks is carried out using the benchmark dataset. The goal is to identify five lightweight EDN configurations that yield the highest evaluation scores in the benchmark dataset. These five Students will then be used to study SDA and compare it to KD and DML. The investigation is structured in four steps described below:

\begin{enumerate}[Step 1.]
    \item \textit{Hyperparameter search}: A CNN encoder is randomly selected and integrated into the EDN of U-Net for this search. Apple's MobileOne-s1 CNN is selected as a random choice. The EDN is then trained with different hyperparameters such as the learning rate, converging epoch, loss functions, and optimisers are experimented. In particular, five popular optimiser algorithms and nine loss functions are studied. The optimisers are supported with a scheduler to optimise the learning rate and the networks are saved at the converging epoch.

    \item \textit{Comparison of EDNs}: Other state-of-the-art EDNs are then studied with the MobileOne-s1 CNN and the identified hyperparameters from Step 1. Due to their popularity eight EDNs are studied: U-Net \citep{ronneberger2015u}, U-Net++ \citep{zhou2019unet++}, U-Net3+ \citep{huang2020unet}, LinkNet \citep{chaurasia2017linknet}, PSPNet \citep{zhao2017pyramid}, FPN \citep{lin2017feature}, DeepLabv3+ \citep{chen2018encoder}, and MANet \citep{fan2020manet} as potential architectures for the Teacher and Student networks. An EDN with the best trade-off between the network parameters (size) and the evaluation scores is selected.

    \item \textit{Student search}: Next, we investigate and benchmark 43 lightweight CNNs available in the computer vision domain, which can be used to configure the lightweight EDNs needed to form a Student network. We identified 43 such CNNs from 16 families $-$ ResNet, DenseNet, MobileNet, MnasNet, EfficientNet, SK-ResNet, Dual path network (DPN), ResNeSt, GERNet, MobileOne, HRNet, MobileViT, FBNet, HardCoRe-NAS, MixNet, and TinyNet. We explain the CNNs with references at \href{https://github.com/bipulneupane/TeacherStudentBuildingExtractor}{https://github.com/bipulneupane/TeacherStudentBuild\\ingExtractor}. These CNNs are integrated into the EDN identified in Step 2 and trained with the hyperparameters from Step 1. The configurations that provide the optimal trade-off between network parameters, loss, training time, and evaluation scores are selected as the Students.

    \item \textit{Teacher search}: For SDA, the EDN configuration of the Student and Teacher remains the same. However, in KD and DML, the Teacher is a large immovable network used to distil smaller Students. Therefore, in Step 4, the larger version of the Student's CNN is studied as the Teacher. Apart from this configuration, we also study U-VGG19 as a Teacher, because of its high scores in the benchmarks \citep{neupane2024cnns}.
\end{enumerate}

\subsection{Training environment}\label{sec:meth:framework}
All DL networks are wrapped in the \textit{PyTorch} framework using \textit{Segmentation Models Pytorch} library \citep{Iakubovskii2019}. The Teacher and Students are trained up to 50 and 200 epochs respectively. The learning rate is reduced upon a plateau of intersection over union (IoU) metric by a factor of 0.1 with patience of 10 epochs. A \textit{sigmoid} function classifies the final binary outputs. Four accuracy metrics are used for evaluation: precision (P), recall (R), IoU, and F1 score. The mathematical notation of the metrics is from \citep{neupane2024cnns}.

\section{Results and Discussion}

\subsection{Preliminary investigation of hyperparameters, EDNs, Students and Teacher}\label{sec:res:prelim}

\subsubsection*{Step 1: Hyperparameter search}\label{sec:res:hyper}
The optimal hyperparameters were searched with a randomly selected MobileOne-s1 CNN encoder for U-Net EDN (abbr. U-MobileOne-s1). Then using the loss function of \textit{total loss}, five optimisers $-$ Adam, stochastic gradient descent (SGD), RMSprop, Adadelta, and Nadam $-$ were compared to find the optimiser that yields the highest evaluation scores. The \textit{total loss} was formulated as the sum of \textit{dice loss} and \textit{binary focal loss}. The \textit{RMSProp} optimiser produced the highest scores for our experimental settings with a trade-off between loss and faster-converging epoch. Then using \textit{RMSProp}, nine loss functions $-$ \textit{binary cross entropy (BCE), BCE dice loss, BCE Jaccard loss, binary focal dice loss, binary focal Jaccard loss, Jaccard loss, dice loss, binary focal loss}, and \textit{total loss} $-$ were further compared. A combination of \textit{RMSProp} optimiser and \textit{dice loss} yielded the highest scores as shown in Table \ref{tab:hyperparam}. \textit{Adam} was marginally close to \textit{RMSProp} optimiser. \textit{Jaccard loss} produced the same IoU and F1 with a faster-converging epoch compared to \textit{dice loss} but yielded a lower loss. The learning rate of 0.0001 produced the highest scores in most experiments.

\begin{table}[ht]
\caption{Hyperparameter search using a U-Net with randomly chosen lightweight CNN of MobileOne-s1. Five optimisers are tested with a Total loss (Dice loss + binary focal loss). The optimiser with the highest F1 score is then used to test the nine loss functions. Columns `ms/it' (millisecond/iteration) and `Ep' report training time and converging epoch. The highest scores are highlighted in bold.}
\label{tab:hyperparam}
\addtolength{\tabcolsep}{-0.25em}
\centering
\footnotesize
\begin{tabular}{lccccccc}
\hline
 & Loss & P & R & IoU & F1 & ms/it & Ep. \\ \hline
\multicolumn{8}{c}{Optimisers (with Total Loss)} \\ \hline
Adam & \textbf{0.177} & \textbf{0.959} & 0.963 & 0.926 & 0.960 & 376 & 26 \\
SGD & 0.338 & 0.889 & \textbf{0.990} & 0.882 & 0.935 & \textbf{200} & 28 \\
RMSProp & \textbf{0.177} & 0.958 & 0.964 & \textbf{0.927} & \textbf{0.961} & 294 & 22 \\
Adadelta & 0.444 & 0.927 & 0.784 & 0.744 & 0.845 & 420 & 29 \\
Nadam & 0.179 & 0.957 & 0.965 & \textbf{0.927} & \textbf{0.961} & 395 & \textbf{20} \\ \hline
\multicolumn{8}{c}{Loss Functions (with RMSProp optimiser)} \\ \hline
BCE & 0.200 & 0.950 & 0.973 & 0.927 & 0.961 & 658 & 26 \\
BCE Dice & 0.243 & 0.949 & 0.975 & \textbf{0.928} & 0.961 & 658 & 15 \\
BCE Jaccard & 0.285 & 0.946 & \textbf{0.978} & 0.927 & 0.961 & 312 & 14 \\
Binary Focal Dice & 0.191 & 0.951 & 0.966 & 0.922 & 0.958 & 324 & \textbf{7} \\
Binary Focal Jaccard & 0.217 & 0.955 & 0.967 & 0.926 & 0.960 & 292 & 26 \\
Jaccard & 0.072 & 0.953 & 0.972 & \textbf{0.928} & \textbf{0.962} & \textbf{282} & 18 \\
Dice & \textbf{0.039} & 0.948 & 0.977 & \textbf{0.928} & \textbf{0.962} & 314 & 20 \\
Binary Focal & 0.120 & \textbf{0.975} & 0.896 & 0.876 & 0.932 & 292 & 30 \\
Total Loss & 0.177 & 0.958 & 0.964 & 0.927 & 0.961 & 294 & 22 \\ \hline
\end{tabular}
\end{table}

The hyperparameter search identified \textit{dice loss} ($L_{dice}$ as the best loss function, which is further used for pre-training the Teacher and adapting Students in SDA. A dice loss \citep{sudre2017generalised} can be expressed as:

    \begin{equation}\label{eqn:dicelossteacher}
        L_{dice} = 1 - \frac{2 y_j p_j + 1} {y_j + p_j + 1} 
    \end{equation}

\noindent where $y_j$ and $p_j$ represent the GT and prediction respectively for the $j^{th}$ pixel. The smooth value of 1 is added in the numerator and denominator to make sure that the function is defined in the case of $y_j = p_j = 0$, which is called an edge case scenario. The product of $y_j$ and $p_j$ represents the intersection between the GT and prediction.

For the KD and DML, we set up the weighted sum of the Student's \textit{dice loss} and a \textit{distillation loss} to effectively reduce the differences in multi-scale feature maps between the Teacher and Students. The loss function setup for KD and DML is described in the supplementary material.

\subsubsection*{Step 2: Comparison of EDNs}\label{sec:res:edns}
With the best hyperparameters found, we further used the U-MobileOne-s1 with RMSProp optimiser and \textit{dice loss} to investigate the other EDNs. The results are shown in the first half of Table \ref{tab:ednencoder}. Results indicated that U-Net++ achieves the highest scores (0.964 F1), closely followed by U-Net, which offered faster training and lower network parameters. Compared to U-Net++, U-Net provided the best trade-off between network parameters (9.1M vs. 12.3M) and evaluation scores with a marginally lower F1 of 0.962 and less complexity in network architecture. Opting for U-Net's slightly lower evaluation scores, we prioritised lower parameters and faster training for small deployable Students. Fewer parameters mean smaller model size and faster training time, which is essential for Students in knowledge transfer. Among the other EDNs, MANet matched U-Net in IoU and F1 but with nearly 6x more parameters. LinkNet and DeepLabv3+ were lightweight but yielded lower scores. FPN and PSPNet, despite faster training, over-fit on the dataset. U-Net is therefore the best EDN configuration for Students.

\begin{table}[!ht]
\caption{Search of EDN and CNN encoders for Student networks. Eight EDNs are first tested with a randomly chosen MobileOne-s1 CNN to identify the EDN with the best trade-off between network parameters and evaluation scores. The highest scores among the EDNs are highlighted in bold. This EDN is then used to test and benchmark the 43 identified lightweight CNNs. Network parameters (Par.) are shown in  Millions. The top 3 scores for the lightweight CNNs are underlined and ranked in order from 1 to 3 with a postscript.}
\label{tab:ednencoder}
\addtolength{\tabcolsep}{-0.25em}
\centering
\footnotesize
\begin{tabular}{lccccccc}
\hline
 & Par. & Loss & P & R & IoU & F1 & ms/it \\ \hline
\multicolumn{8}{c}{EDN with MobileOne-s1 encoder} \\ \hline
U-Net & 9.1 & 0.039 & 0.948 & 0.977 & 0.928 & 0.962 & 314 \\
DeepLabv3+ & 5.7 & 0.044 & 0.937 & \textbf{0.979} & 0.919 & 0.956 & 287 \\
LinkNet & 6.2 & 0.042 & 0.945 & 0.973 & 0.922 & 0.958 & 326 \\
MANet & 53.3 & 0.039 & 0.948 & 0.976 & 0.928 & 0.961 & 376 \\
U-Net++ & 12.3 & \textbf{0.036} & 0.954 & 0.975 & \textbf{0.933} & \textbf{0.964} & 395 \\
FPN & 5.7 & 0.062 & 0.887 & 1.000 & 0.887 & 0.938 & 309 \\
PSPNet & \textbf{3.8} & 0.062 & 0.887 & 1.000 & 0.887 & 0.938 & \textbf{168} \\ \hline
\multicolumn{8}{c}{U-Net (best trade-off of Par. and F1) with lightweight CNNs as encoder} \\ \hline
ResNet-18 & 14.3 & 0.038 & 0.949 & \underline{0.977}$^3$ & 0.929 & 0.962 & 109 \\
DenseNet-121 & 13.6 & \underline{0.037}$^3$ & 0.950 & \underline{0.977}$^3$ & \underline{0.931}$^3$ & \underline{0.963}$^3$ & 629 \\
SE-ResNet-18 & 14.4 & 0.039 & 0.950 & 0.974 & 0.927 & 0.961 & \underline{151}$^1$ \\
MobileNet-v2 & 6.6 & 0.039 & 0.951 & 0.972 & 0.927 & 0.961 & 201 \\
MobileNetv3-s & 3.6 & 0.044 & 0.940 & 0.974 & 0.918 & 0.956 & 281 \\
MobileNetv3-l & 6.7 & 0.043 & 0.946 & 0.970 & 0.921 & 0.957 & 316 \\
Eff.NetB0 & 6.3 & 0.038 & \underline{0.952}$^3$ & 0.973 & 0.929 & 0.962 & 304 \\
Eff.NetB1 & 8.8 & 0.038 & 0.950 & 0.976 & 0.930 & 0.962 & 461 \\
Eff.NetB2 & 10.0 & \underline{0.037}$^3$ & \underline{0.956}$^1$ & 0.971 & \underline{0.931}$^3$ & \underline{0.963}$^3$ & 459 \\
Eff.NetB3 & 13.2 & \underline{0.037}$^3$ & 0.951 & \underline{0.977}$^3$ & 0.930 & \underline{0.963}$^3$ & 485 \\
Eff.Net-lite0 & 5.6 & 0.039 & 0.950 & 0.974 & 0.928 & 0.962 & 187 \\
Eff.Net-lite1 & 6.4 & 0.039 & 0.948 & 0.976 & 0.928 & 0.961 & 233 \\
Eff.Net-lite2 & 7.2 & 0.039 & 0.948 & 0.975 & 0.927 & 0.961 & 229 \\
Eff.Net-lite3 & 9.4 & 0.038 & 0.948 & \underline{0.978}$^2$ & 0.929 & 0.962 & 258 \\
Eff.Net-lite4 & 14.4 & 0.040 & 0.945 & \underline{0.978}$^2$ & 0.926 & 0.960 & 313 \\
Eff.Netv2B0 & 7.6 & 0.038 & 0.951 & 0.975 & 0.929 & 0.962 & 284 \\
Eff.Netv2B1 & 8.6 & \underline{0.037}$^3$ & \underline{0.953}$^2$ & 0.975 & \underline{0.931}$^3$ & \underline{0.963}$^3$ & 334 \\
Eff.Netv2B2 & 10.4 & \underline{0.037}$^3$ & 0.951 & \underline{0.977}$^3$ & \underline{0.931}$^3$ & \underline{0.963}$^3$ & 347 \\
Eff.Netv2B3 & 14.6 & \underline{0.035}$^1$ & \underline{0.956}$^1$ & 0.975 & \underline{0.934}$^1$ & \underline{0.965}$^1$ & 408 \\
SK-ResNet-18 & 14.6 & 0.038 & 0.948 & \underline{0.978}$^2$ & 0.929 & 0.962 & 201 \\
DPN-68 & 17.0 & 0.041 & 0.945 & 0.975 & 0.924 & 0.959 & 493 \\
ResNeSt-14 & 17.6 & \underline{0.036}$^2$ & \underline{0.952}$^3$ & \underline{0.977}$^3$ & \underline{0.932}$^2$ & \underline{0.964}$^2$ & \underline{159}$^2$ \\
GERNet-s & 12.8 & 0.040 & 0.947 & 0.974 & 0.925 & 0.960 & \underline{180}$^3$ \\
MobileOne-s0 & 8.6 & 0.038 & 0.948 & \underline{0.979}$^1$ & 0.930 & 0.963 & 610 \\
MobileOne-s1 & 9.1 & 0.039 & 0.950 & 0.975 & 0.928 & 0.962 & 287 \\
MobileOne-s2 & 13.6 & 0.039 & 0.949 & 0.975 & 0.928 & 0.961 & 296 \\
MobileOne-s3 & 16.2 & 0.039 & 0.950 & 0.973 & 0.927 & 0.961 & 293 \\
HRNet-18 & 16.1 & 0.038 & \underline{0.952}$^3$ & 0.974 & 0.929 & 0.962 & 870 \\
MNASNet-s & \underline{2.2}$^1$ & 0.043 & 0.938 & \underline{0.978}$^2$ & 0.920 & 0.957 & 214 \\
MobileViT-s & 8.0 & 0.040 & 0.946 & 0.975 & 0.925 & 0.960 & 289 \\
MobileViT-xs & 4.3 & 0.040 & 0.947 & 0.975 & 0.926 & 0.960 & 294 \\
MobileViT-xxs & \underline{3.1}$^3$ & 0.042 & 0.943 & 0.975 & 0.922 & 0.958 & 299 \\
FBNet-c100 & 5.2 & 0.041 & 0.947 & 0.973 & 0.924 & 0.959 & 229 \\
FBNet-v3b & 8.6 & 0.042 & 0.943 & 0.974 & 0.921 & 0.958 & 690 \\
FBNet-v3d & 10.3 & 0.041 & 0.943 & \underline{0.978}$^2$ & 0.925 & 0.960 & 769 \\
FBNet-v3g & 16.8 & 0.040 & 0.945 & 0.976 & 0.925 & 0.960 & 917 \\
HardCoRe-NAS-a & 6.5 & 0.042 & 0.940 & \underline{0.979}$^1$ & 0.922 & 0.958 & 283 \\
HardCoRe-NAS-f & 9.4 & 0.043 & 0.937 & 0.980 & 0.920 & 0.957 & 552 \\
MixNet-s & 4.3 & 0.044 & 0.943 & 0.970 & 0.918 & 0.956 & 429 \\
MixNet-m & 5.2 & 0.043 & 0.943 & 0.974 & 0.921 & 0.957 & 498 \\
MixNet-l & 7.6 & 0.041 & 0.948 & 0.972 & 0.924 & 0.959 & 532 \\
TinyNet-a & 6.7 & 0.042 & 0.944 & 0.975 & 0.923 & 0.959 & 296 \\
TinyNet-e & \underline{2.3}$^2$ & 0.045 & 0.934 & \underline{0.979}$^1$ & 0.916 & 0.955 & 182 \\ \hline
\end{tabular}
\end{table}

\subsubsection*{Step 3: Student search}\label{sec:res:student}
\begin{figure}[!ht]
  \centering
  \includegraphics[width=1.0\linewidth]{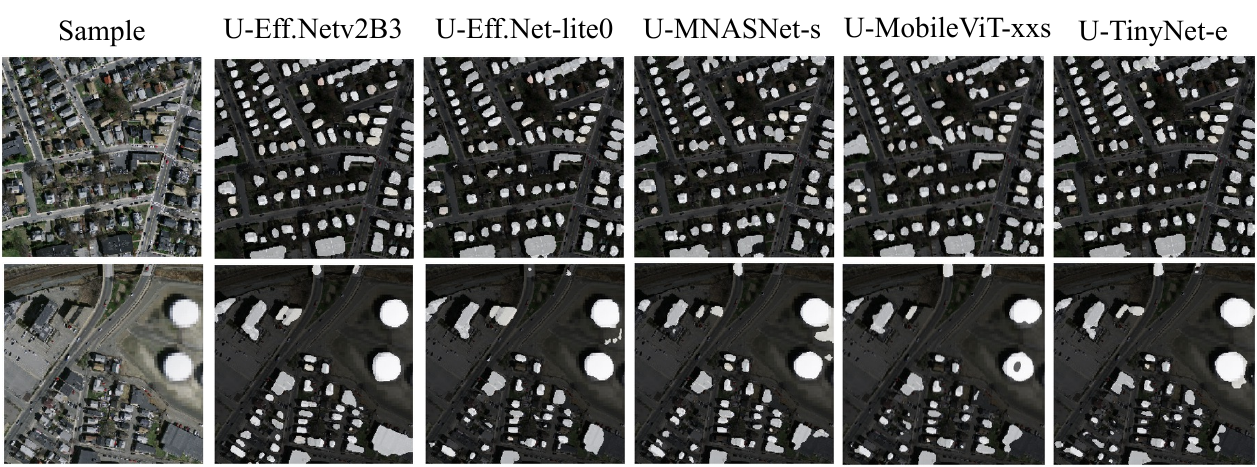}
  \caption{Segmentation results of the selected lightweight Students on the subset of Massachusetts building dataset. U-EfficientNetv2B3 produced the highest F1 score of 0.965. Other Students produced lower scores but with trade-offs between network parameters, training time, loss, and evaluation scores as seen in Table \ref{tab:ednencoder}.}
      \label{fig:masssmallres}
      \end{figure}

After identifying U-Net as the best-performing EDN, the 43 lightweight CNNs identified from the computer vision domain were integrated into U-Net for the Student search. Thus developed U-Net configurations of the 43 CNNs are shown in the second half of Table \ref{tab:ednencoder}. The high scorer was U-EfficientNetv2B3 (0.965 F1), U-EfficientNet-lite0 provided the best trade-off between evaluation scores and network parameters, and the lightest Students were U-MNASNet-s (2.2M), U-TinyNet-e (2.3M), and U-MobileViT-xxs (3.1M). These EDNs were therefore selected as the top 5 Students for the rest of the study. Figure \ref{fig:masssmallres} shows the segmentation outputs from these five Students on two random samples from the validation subset of the Massachusetts building dataset. The trend showed that the segmentation performance decreases with the number of network parameters.

\subsubsection*{Step 4: Teacher search}\label{sec:res:teacher}
\begin{figure*}[!htb]
  \centering
  \includegraphics[width=\linewidth]{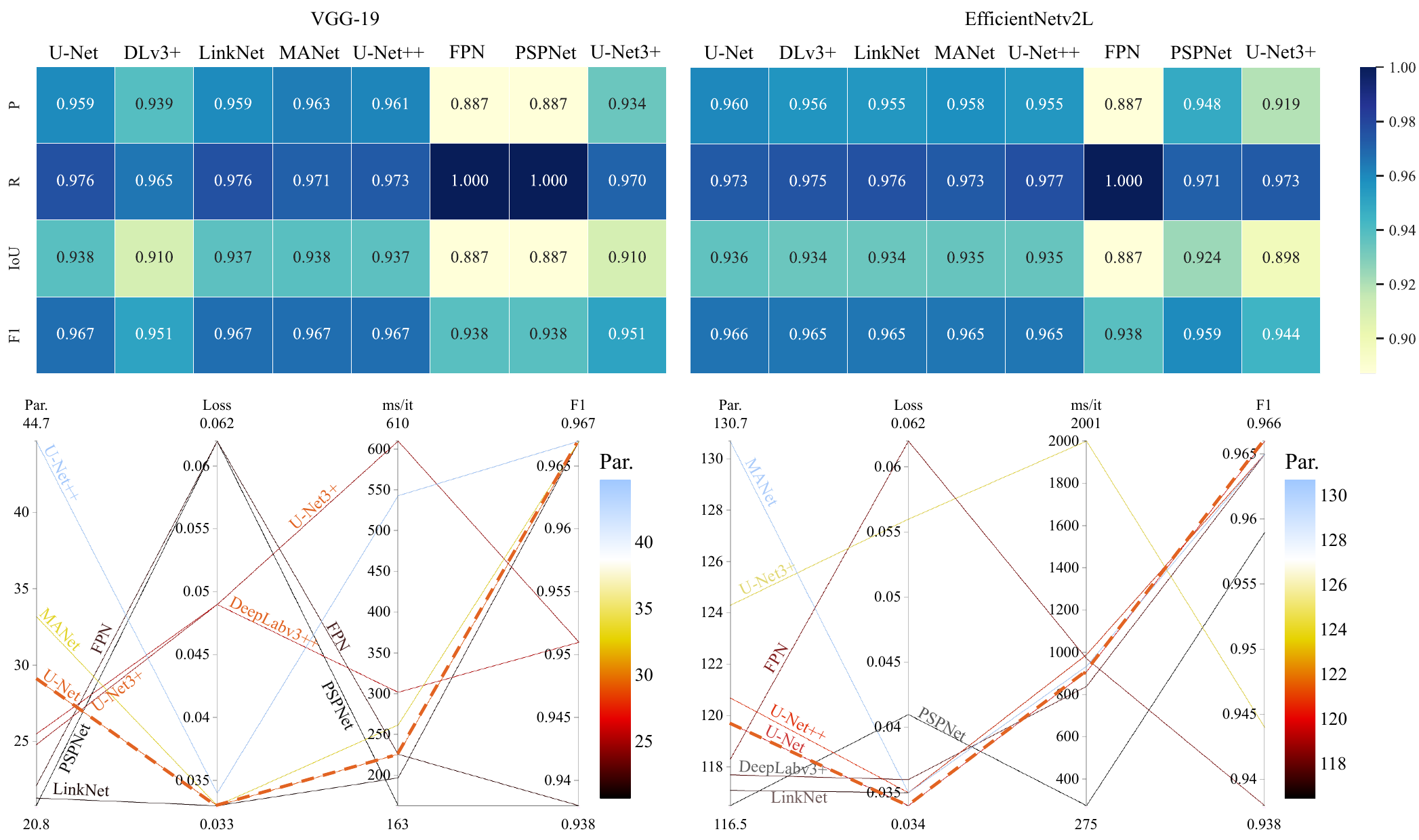}
  \caption{Teacher search with VGG-19 (left) and EfficientNetv2B3 (right) as CNNs for the state-of-the-art EDNs. The heat maps on top compare the evaluation scores of the EDNs. The parallel coordinate plots on the bottom show the trade-off between network parameters (Par.), loss, training time (ms/it), and F1 score. The network with the dashed line (U-Net) provides the best trade-off among the variables.}
      \label{fig:edncompare}
      \end{figure*}

For SDA, the Teacher and Student constitute the same EDN network. For KD and DML, the best Teacher was identified from the comparison of VGG19 (top scorer from \citep{neupane2024cnns}) and EfficientNetv2L (largest version of the best-performing CNN of EfficientNetv2B3) with eight EDNs. The heatmaps and the parallel coordinate plots in Figure \ref{fig:edncompare} provide minute details to identify the most optimal trade-off from the networks. With VGG-19, three EDNs of U-Net, LinkNet, and U-Net++ achieved the highest average of the four accuracy metrics (0.960) and an F1 of 0.967. However, with EfficientNetv2L, only U-Net yielded the highest F1 score, higher than DeepLabv3+, LinkNet, MANet, and U-Net++. LinkNet provided the best trade-off with VGG-19, while U-Net provided that with the highest F1 and lower parameters with both CNNs. Therefore, U-VGG19 is the Teacher for KD and DML.

\subsection{Baseline scores for comparison}\label{sec:res:quantitativeanalysis}
With the Teacher and Students identified, we trained them on the T and S datasets alone and produced the validation results on Ev to prepare the baseline scores for SDA, KD, and DML. The scores were produced in different settings of varying \textit{train} and validation (\textit{val}) subsets of T, S, and Ev datasets.

\textit{Teacher (U-VGG19)'s Baseline Scores:} were produced in three settings $-$ T-T, T-S, and T-Ev $-$ and reported in Table \ref{tab:teacherbaseline}. T-T refers to training and validation on T, T-S refers to training on T and validation on S, and T-Ev refers to training on T and validation on Ev. From the difference in T-T and T-S settings, it was evident that the Teacher's ability to generalise the validation samples from S was influenced by the domain shift between T and S. This domain shift can be attributed to two primary factors: the presence of unfamiliar complex urban high-rise buildings in S and the contrast between precision of ortho-rectification between the samples of T and S. However, in the T-Ev setting, the Teacher demonstrated its capacity to generalise complex high-rise buildings in Ev, even though it was trained on misaligned labels in T. The complexity of the clean data S and Ev was prominent in the experimental results.

\begin{table}[!ht]
\caption{Performance of Teacher U-VGG19 trained on T and validated on three datasets (T, S, and Ev).}
\label{tab:teacherbaseline}
\centering
\footnotesize
\begin{tabular}{lcccccc}
\hline
Setting & Loss & P & R & IoU & F1 & ms/it \\ \hline
T-T & 0.080 & 0.929 & 0.917 & 0.860 & 0.920 & 235 \\
T-S & 0.186 & 0.770 & 0.894 & 0.713 & 0.814 & 40 \\
T-Ev & 0.169 & 0.877 & 0.814 & 0.746 & 0.831 & 40 \\ \hline
\end{tabular}
\end{table}

\textit{Student's Baseline Scores (without knowledge transfer)} were studied in S-S and S-Ev settings. S-S means training on S and validation on S; S-Ev means training on S and validation on Ev. Table \ref{tab:maincomparison} reports their baseline scores along with the performance of the three knowledge transfer methods. U-VGG19 (0.893 F1) and U-EfficientNetv2B3 (0.819 F1) score the highest when validated on S and Ev respectively. The performance of Students after SDA and other knowledge transfer methods will be evaluated against these baseline scores.

\begin{table*}[!ht]
\caption{Table to report the comparison of the knowledge transfer methods. The table is divided into four parts: Student's Baseline Sores (trained alone on S), SDA, KD, and DML. The methods from each part are evaluated in two settings S-S and S-Ev. The KD and DML are based on the Teacher U-VGG19. The Teacher and Students are identified from the preliminary investigation. DML is performed between six pairs of Students. Column `ms/it' and `Par. Red. (\%)' respectively report the training time and the percentage of reduced network parameters.}
\label{tab:maincomparison}
\centering
\footnotesize
\begin{tabular}{lccccccccccccc}
\hline
\multicolumn{1}{l|}{Network} & Loss & P & R & IoU & F1 & \multicolumn{1}{c|}{ms/it} & Loss & P & R & IoU & F1 & \multicolumn{1}{c|}{ms/it} & Par. Red.(\%) \\ \hline
\multicolumn{14}{c}{Student's Baseline Scores (Trained alone on S)} \\ \hline
\multicolumn{1}{l|}{} & \multicolumn{6}{c|}{S-S: Trained and Validated on S} & \multicolumn{6}{c|}{S-Ev: Trained on S and validated on Ev} & \multicolumn{1}{l}{} \\ \hline
\multicolumn{1}{l|}{U-VGG19} & \textbf{0.107} & \textbf{0.890} & \textbf{0.913} & \textbf{0.824} & \textbf{0.893} & \multicolumn{1}{c|}{224} & 0.215 & \textbf{0.866} & 0.741 & 0.676 & 0.785 & \multicolumn{1}{c|}{\textbf{41}} & - \\
\multicolumn{1}{l|}{U-Eff.Net-lite0} & 0.155 & 0.838 & 0.877 & 0.754 & 0.845 & \multicolumn{1}{c|}{190} & 0.195 & 0.816 & 0.814 & 0.687 & 0.805 & \multicolumn{1}{c|}{75} & - \\
\multicolumn{1}{l|}{U-Eff.Netv2B3} & 0.145 & 0.840 & 0.901 & 0.769 & 0.855 & \multicolumn{1}{c|}{392} & \textbf{0.181} & 0.838 & \textbf{0.819} & \textbf{0.708} & \textbf{0.819} & \multicolumn{1}{c|}{96} & - \\
\multicolumn{1}{l|}{U-MNASNet-s} & 0.184 & 0.797 & 0.872 & 0.710 & 0.816 & \multicolumn{1}{c|}{219} & 0.235 & 0.784 & 0.773 & 0.639 & 0.765 & \multicolumn{1}{c|}{56} & - \\
\multicolumn{1}{l|}{U-MobileViT-xxs} & 0.181 & 0.803 & 0.872 & 0.717 & 0.819 & \multicolumn{1}{c|}{296} & 0.233 & 0.814 & 0.748 & 0.634 & 0.767 & \multicolumn{1}{c|}{154} & - \\
\multicolumn{1}{l|}{U-TinyNet-e} & 0.174 & 0.795 & 0.896 & 0.727 & 0.826 & \multicolumn{1}{c|}{\textbf{181}} & 0.221 & 0.808 & 0.776 & 0.648 & 0.779 & \multicolumn{1}{c|}{56} & - \\ \hline
\multicolumn{14}{c}{Supervised domain adaptation (SDA)} \\ \hline
 & \multicolumn{6}{c|}{S-S: SDA from T to S, and validated on S} & \multicolumn{6}{c|}{S-Ev: SDA from T to S, and validated on Ev} & \multicolumn{1}{l}{} \\ \hline
\multicolumn{1}{l|}{U-VGG19} & \textbf{0.098} & \textbf{0.906} & \textbf{0.914} & \textbf{0.838} & \textbf{0.903} & \multicolumn{1}{c|}{231} & 0.173 & \textbf{0.917} & 0.784 & 0.737 & 0.827 & \multicolumn{1}{c|}{48} & - \\
\multicolumn{1}{l|}{U-Eff.Net-lite0} & 0.138 & 0.859 & 0.890 & 0.778 & 0.862 & \multicolumn{1}{c|}{194} & 0.197 & 0.873 & 0.763 & 0.688 & 0.803 & \multicolumn{1}{c|}{103} & - \\
\multicolumn{1}{l|}{U-Eff.Netv2B3} & 0.120 & 0.869 & \textbf{0.914} & 0.805 & 0.880 & \multicolumn{1}{c|}{386} & \textbf{0.153} & 0.874 & \textbf{0.837} & \textbf{0.752} & \textbf{0.847} & \multicolumn{1}{c|}{95} & - \\
\multicolumn{1}{l|}{U-MNASNet-s} & 0.151 & 0.828 & 0.900 & 0.759 & 0.849 & \multicolumn{1}{c|}{215} & 0.205 & 0.844 & 0.772 & 0.690 & 0.795 & \multicolumn{1}{c|}{\textbf{55}} & - \\
\multicolumn{1}{l|}{U-MobileViT-xxs} & 0.145 & 0.846 & 0.890 & 0.769 & 0.855 & \multicolumn{1}{c|}{294} & 0.191 & 0.848 & 0.793 & 0.694 & 0.809 & \multicolumn{1}{c|}{149} & - \\
\multicolumn{1}{l|}{U-TinyNet-e} & 0.160 & 0.831 & 0.882 & 0.748 & 0.840 & \multicolumn{1}{c|}{\textbf{181}} & 0.222 & 0.844 & 0.741 & 0.653 & 0.779 & \multicolumn{1}{c|}{56} & - \\ \hline
\multicolumn{14}{c}{Knowledge Distillation (KD) with Teacher U-VGG19} \\ \hline
\multicolumn{1}{l|}{} & \multicolumn{6}{c|}{S-S: Distilled on S, validated on S} & \multicolumn{6}{c|}{S-Ev: Distilled on S, validated on Ev} & \multicolumn{1}{l}{} \\ \hline
\multicolumn{1}{l|}{U-VGG19} & \textbf{0.099} & 0.907 & \textbf{0.895} & \textbf{0.820} & \textbf{0.901} & \multicolumn{1}{c|}{410} & \textbf{0.184} & \textbf{0.877} & 0.772 & \textbf{0.707} & \textbf{0.812} & \multicolumn{1}{c|}{40} & 0 \\
\multicolumn{1}{l|}{U-Eff.Net-lite0} & 0.265 & \textbf{0.944} & 0.602 & 0.581 & 0.735 & \multicolumn{1}{c|}{306} & 0.232 & 0.794 & 0.802 & 0.656 & 0.784 & \multicolumn{1}{c|}{43} & 82 \\
\multicolumn{1}{l|}{U-Eff.Netv2B3} & 0.149 & 0.895 & 0.810 & 0.740 & 0.851 & \multicolumn{1}{c|}{467} & 0.219 & 0.844 & 0.735 & 0.650 & 0.774 & \multicolumn{1}{c|}{77} & 50 \\
\multicolumn{1}{l|}{U-MNASNet-s} & 0.199 & 0.869 & 0.744 & 0.669 & 0.801 & \multicolumn{1}{c|}{278} & 0.223 & 0.756 & \textbf{0.842} & 0.654 & 0.781 & \multicolumn{1}{c|}{52} & 93 \\
\multicolumn{1}{l|}{U-MobileViT-xxs} & 0.214 & 0.811 & 0.763 & 0.648 & 0.786 & \multicolumn{1}{c|}{346} & 0.320 & 0.840 & 0.610 & 0.538 & 0.688 & \multicolumn{1}{c|}{93} & 89 \\
\multicolumn{1}{l|}{U-TinyNet-e} & 0.179 & 0.860 & 0.785 & 0.696 & 0.821 & \multicolumn{1}{c|}{\textbf{315}} & 0.294 & 0.826 & 0.638 & 0.554 & 0.702 & \multicolumn{1}{c|}{\textbf{46}} & 92 \\ \hline
\multicolumn{14}{c}{Deep Mutual Learning (DML) of six Student pairs that are distilled from Teacher U-VGG19} \\ \hline
\multicolumn{1}{l|}{} & \multicolumn{6}{c|}{S-S: Distilled on S, validated on S} & \multicolumn{6}{c|}{S-Ev: Distilled on S, validated on Ev} & \multicolumn{1}{l}{} \\ \hline
\multicolumn{1}{l|}{U-VGG19} & 0.118 & 0.912 & 0.873 & 0.806 & 0.882 & \multicolumn{1}{c|}{\multirow{2}{*}{862}} & 0.201 & 0.888 & 0.739 & 0.685 & 0.793 & \multicolumn{1}{c|}{\textbf{41}} & 0 \\
\multicolumn{1}{l|}{U-Eff.Netv2B3} & 0.168 & 0.881 & 0.819 & 0.731 & 0.832 & \multicolumn{1}{c|}{} & \textbf{0.192} & 0.829 & 0.810 & \textbf{0.696} & \textbf{0.810} & \multicolumn{1}{c|}{78} & 82 \\ \hline
\multicolumn{1}{l|}{U-VGG19} & 0.121 & \textbf{0.913} & 0.869 & 0.802 & 0.879 & \multicolumn{1}{c|}{\multirow{2}{*}{699}} & 0.196 & \textbf{0.889} & 0.743 & 0.689 & 0.796 & \multicolumn{1}{c|}{\textbf{41}} & 0 \\
\multicolumn{1}{l|}{U-TinyNet-e} & 0.201 & 0.883 & 0.768 & 0.685 & 0.799 & \multicolumn{1}{c|}{} & 0.221 & 0.775 & 0.808 & 0.658 & 0.781 & \multicolumn{1}{c|}{44} & 92 \\ \hline
\multicolumn{1}{l|}{U-VGG19} & \textbf{0.117} & 0.907 & \textbf{0.882} & \textbf{0.808} & \textbf{0.883} & \multicolumn{1}{c|}{\multirow{2}{*}{\textbf{690}}} & 0.204 & 0.883 & 0.731 & 0.678 & 0.784 & \multicolumn{1}{c|}{47} & 0 \\
\multicolumn{1}{l|}{U-MNASNet-s} & 0.191 & 0.880 & 0.785 & 0.701 & 0.809 & \multicolumn{1}{c|}{} & 0.253 & 0.805 & 0.722 & 0.615 & 0.748 & \multicolumn{1}{c|}{50} & 93 \\ \hline
\multicolumn{1}{l|}{U-Eff.Netv2B3} & 0.172 & 0.879 & 0.817 & 0.726 & 0.828 & \multicolumn{1}{c|}{\multirow{2}{*}{775}} & 0.232 & 0.846 & 0.718 & 0.630 & 0.761 & \multicolumn{1}{c|}{75} & 50 \\
\multicolumn{1}{l|}{U-Eff.Net-lite0} & 0.179 & 0.886 & 0.798 & 0.717 & 0.821 & \multicolumn{1}{c|}{} & 0.254 & 0.611 & \textbf{0.999} & 0.610 & 0.746 & \multicolumn{1}{c|}{47} & 82 \\ \hline
\multicolumn{1}{l|}{U-Eff.Netv2B3} & 0.171 & 0.900 & 0.801 & 0.730 & 0.829 & \multicolumn{1}{c|}{\multirow{2}{*}{840}} & 0.313 & 0.864 & 0.597 & 0.541 & 0.685 & \multicolumn{1}{c|}{77} & 50 \\
\multicolumn{1}{l|}{U-TinyNet-e} & 0.196 & 0.881 & 0.776 & 0.693 & 0.804 & \multicolumn{1}{c|}{} & 0.221 & 0.792 & 0.801 & 0.656 & 0.782 & \multicolumn{1}{c|}{47} & 92 \\ \hline
\multicolumn{1}{l|}{U-Eff.Netv2B3} & 0.169 & 0.884 & 0.816 & 0.732 & 0.831 & \multicolumn{1}{c|}{\multirow{2}{*}{800}} & 0.210 & 0.845 & 0.747 & 0.658 & 0.782 & \multicolumn{1}{c|}{76} & 50 \\
\multicolumn{1}{l|}{U-MNASNet-s} & 0.203 & 0.884 & 0.763 & 0.684 & 0.797 & \multicolumn{1}{c|}{} & 0.218 & 0.767 & 0.838 & 0.665 & 0.790 & \multicolumn{1}{c|}{50} & 93 \\ \hline
\end{tabular}
\end{table*}

\subsection{Students adapted with SDA}\label{sec:res:sda}
SDA improved the performance of all Students. Compared to the baseline scores of S-S and S-Ev, all Students in both settings produced higher scores, except U-EfficientNet-lite0 of S-Ev. They also outperformed the Teacher (T-S setting form Table \ref{tab:teacherbaseline}) in S. U-VGG19 and U-EfficientNetv2B3 achieved the highest scores in the settings of S-S (0.903 F1) and S-Ev (0.847 F1) respectively, suggesting the U-EfficientNetv2B3 to be the best Student for SDA in the Ev dataset.

\subsection{Students distilled with KD and DML}\label{sec:res:kd}
\textit{Knowledge Distillation}: The experimental design for KD is described in supplementary material. Compared to the Student's baseline scores, most distilled networks produced lower scores including F1. KD between U-VGG19 and itself produced better results in both S-S and S-Ev settings, but neither higher than SDA nor did it allow model compression. U-MNASNet-s was the only exception that produced higher scores in S-Ev. Among the smaller Students, U-EfficientNetv2B3 and U-EfficientNet-lite0 yielded the highest scores in the settings of S-S (0.851 F1) and S-Ev (0.784 F1) respectively, suggesting U-EfficientNet-lite0 to be the best Student for KD in the Ev dataset with network parameters reduced by 82\% from 29.1M (of U-VGG19) to 5.6M.

\textit{Deep Mutual Learning}: The experimental design for DML is described in the supplementary material. Similar to KD, the experiments were performed between the Teacher of U-VGG19 and the selected Students. Only two Students were distilled at a time because of the computational costs of loading multiple EDNs in the memory. DML of U-MobileViT-xxs was not studied because of its incompetence with KD. In the S-S, all Students failed compared to their baseline scores, and most Student pairs produced lower scores. DML improved some Students in S-Ev, however, reducing the performance of at least one Student for each pair. Two pairs of (U-VGG19, U-TinyNet-e) and (U-EfficientNetv2B3, U-MNASNet-s) increased the F1 scores of both Students, compared to the baseline. The pair of (U-VGG19, U-EfficientNetv2B3) produced the highest scores on average in both settings.

\subsection{SDA vs. baseline scores, KD, and DML}\label{sec:res:sdavskddml}
\begin{figure*}[!ht]
  \centering
  \includegraphics[width=\linewidth]{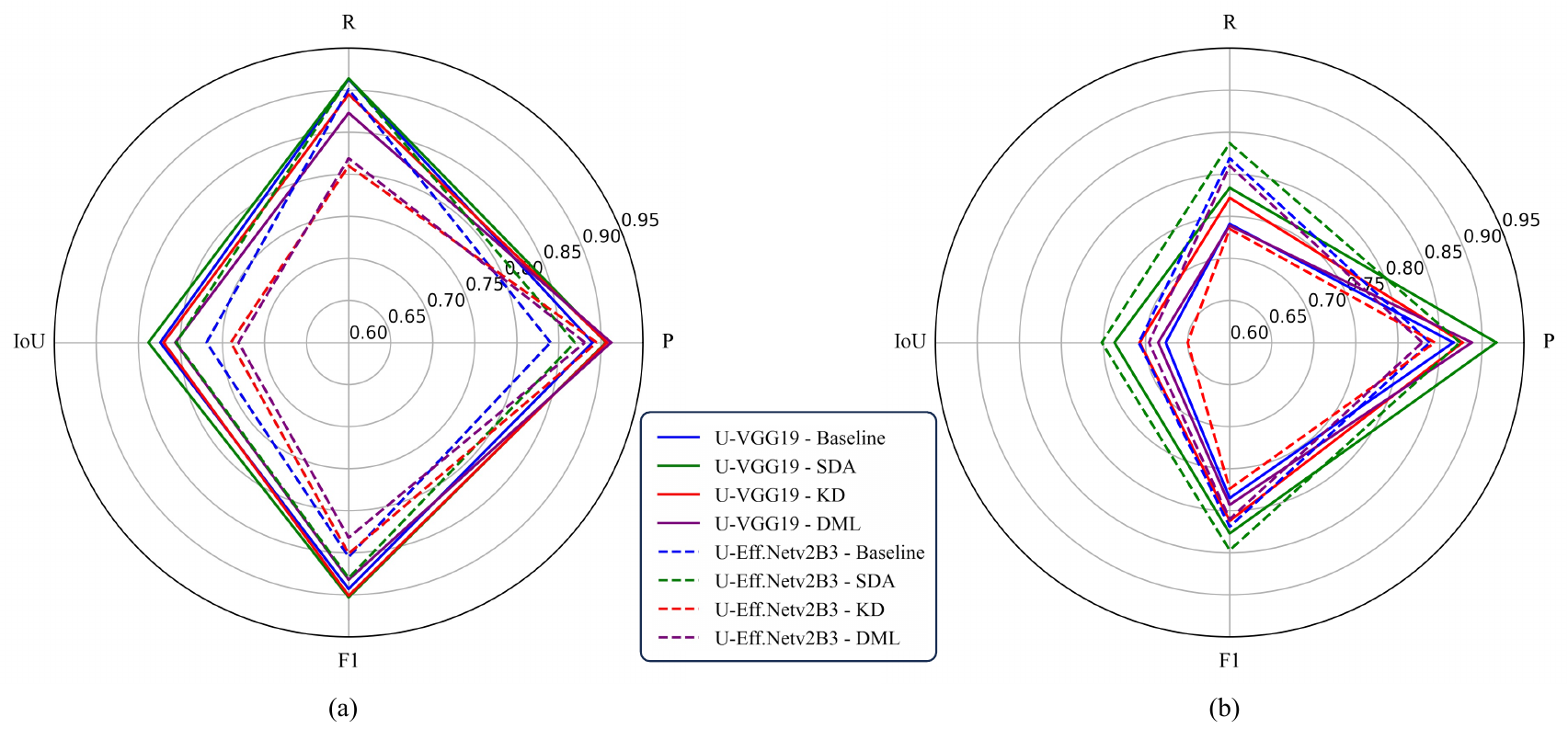}
  \caption{A spider chart to show the summary of results from the three knowledge transfer methods with U-VGG19 and U-EfficientNetv2B3 Students adapted/distilled with U-VGG19 as Teacher in (a) S-S setting and (b) S-Ev setting. In the S-S setting, the SDA of U-VGG19 produced the highest F1 of 0.903, followed by KD (0.901 F1) with the same Student. In the S-Ev setting, the SDA of U-EfficeintNetv2B3 yielded the highest F1 of 0.847, followed by the same network trained alone on S (baseline) with a 0.819 F1.}
      \label{fig:resultsummary}
      \end{figure*}

Finding the best Teacher and Students allowed a fair comparison of SDA against KD and DML. The results from the three knowledge transfer methods can be summarised with common high-scorers in S-S and S-Ev: U-VGG19 and U-EfficientNetv2B3. The summary is reported with a spider chart in Figure \ref{fig:resultsummary}. SDA outperformed the baseline scores and other methods with both U-VGG19 and U-EfficientNetv2B3 in the S-S setting. KD outperformed the baseline scores with U-VGG19. However, using U-VGG19 as the Student did not offer model compression. With U-EfficientNetv2B3, there was a small trade-off of scores with a reduction in 50\% of network parameters (29.1M vs. 14.6M). Compared to baseline F1, the difference was marginal in KD (0.851 vs. 0.855) and DML (0.832 vs. 0.855) with U-EfficientNetv2B3. In the S-Ev setting, SDA outperformed the baseline scores and other methods with both U-VGG19 and U-EfficientNetv2B3. All three methods outperformed the baseline scores with U-VGG19, without offering the model compression. The reduction of F1 with U-EfficientNetv2B3 was 4.5\% (0.774 vs. 0.819) with KD and 0.9\% (0.810 vs. 0.819) with DML, compared to the baseline.

In summary, U-EfficientNetv2B3 was the best Student for SDA but without model compression. KD and DML offered compression, with a reduction in scores. U-EfficientNet-lite0 was the best Student for KD with 82\% fewer network parameters (from 29.1M to 5.6M) compared to the Teacher of U-VGG19. With DML, only a few Student pairs improved the scores of both. The results conclude that SDA significantly minimises the effects of misaligned building labels, compared to KD and DML.

\subsection{Improvements with SDA vs. without SDA}\label{sec:disc:nosda}
\begin{figure*}[!ht]
  \centering
  \includegraphics[width=\linewidth]{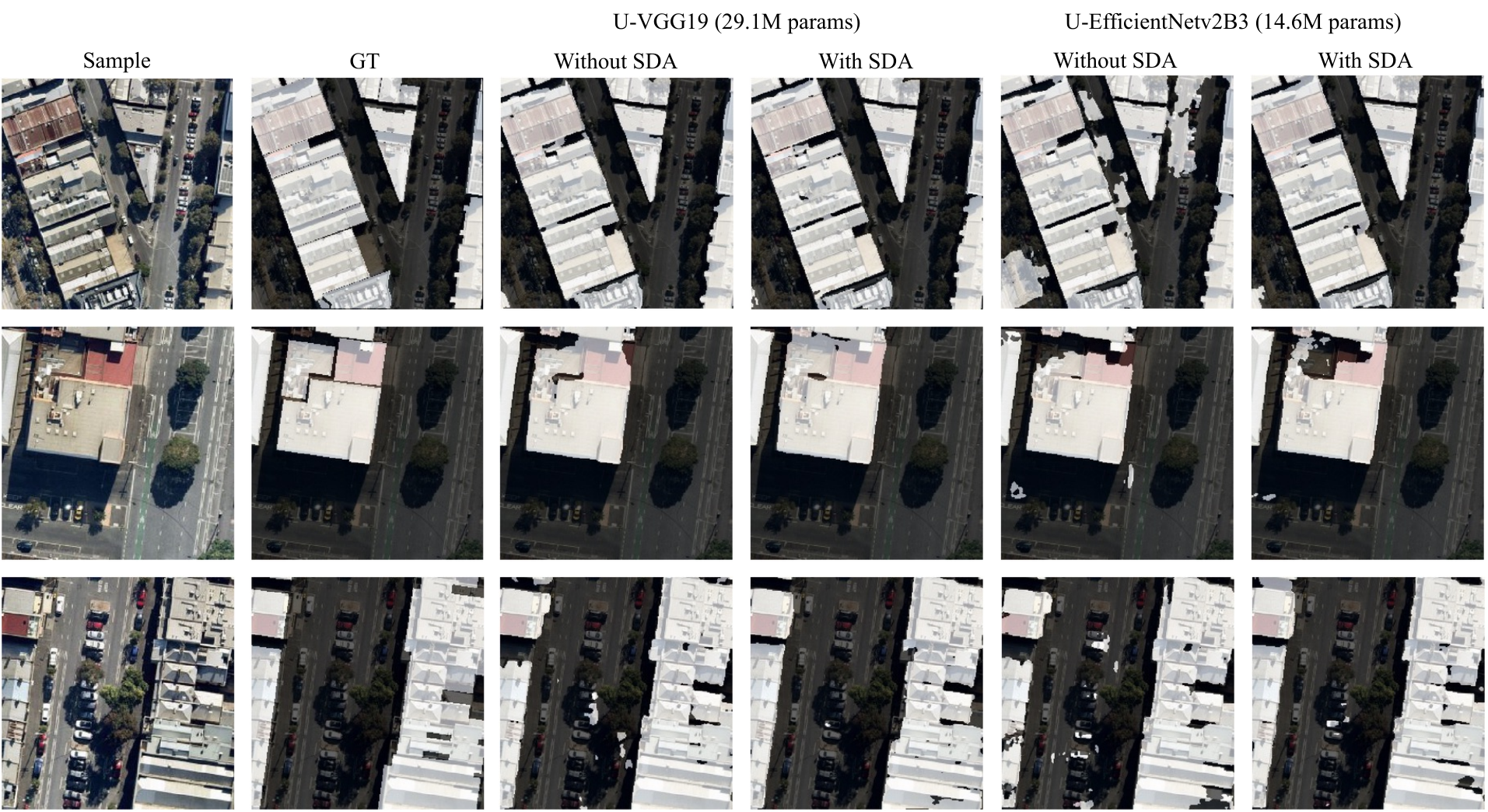}
  \caption{The segmented outputs of the U-VGG19 and U-EfficientNetv2B3 without SDA and with SDA on the data samples of the Ev dataset.}
      \label{fig:sdavsnosda}
      \end{figure*}

Table \ref{tab:sda} shows the percentage gains in Student's IoU and F1 from SDA compared to the Students trained alone without adaptation. SDA improved all Students in IoU (up to 6.1\% in Ev) and most in F1 (up to 4.2\% in Ev). U-VGG19 and U-EfficientNetv2B3 produced the highest scores in S-S and S-Ev settings. The segmented outputs of the two high-scorers without SDA and with SDA are shown in Figure \ref{fig:sdavsnosda}.

\begin{table}[!ht]
\caption{Improvements with SDA vs. without SDA for the Students with different encoders on S and Ev datasets. The gains are shown in percentage and the highest scores are highlighted in bold.}
\label{tab:sda}
\addtolength{\tabcolsep}{-0.3em}
\centering
\scriptsize
\begin{tabular}{lcccc}
\hline
\multirow{2}{*}{Student} & \multicolumn{2}{c}{S-S} & \multicolumn{2}{c}{S-Ev} \\ \cline{2-5} 
 & IoU & F1 & IoU & F1 \\ \hline
U-VGG19 & \textbf{0.838 (+1.4\%)} & \textbf{0.903 (+1.0\%)} & 0.737 (+6.1\%) & 0.827 (+4.2\%) \\
U-Eff.Net-lite0 & 0.778 (+2.4\%) & 0.862 (+1.7\%) & 0.688 (+0.1\%) & 0.803 (-0.2\%) \\
U-Eff.Netv2B3 & 0.805 (+3.6\%) & 0.880 (+2.5\%) & \textbf{0.752 (+4.4\%)} & \textbf{0.847 (+2.8\%)} \\
U-MNASNet-s & 0.759 (+4.9\%) & 0.849 (+3.3\%) & 0.690 (+5.1\%) & 0.795 (+3.0\%) \\
U-MobileViT-xxs & 0.769 (+5.2\%) & 0.855 (+3.6\%) & 0.694 (+6.0\%) & 0.809 (+4.2\%) \\
U-TinyNet-e & 0.748 (+2.1\%) & 0.840 (+1.4\%) & 0.653 (+0.5\%) & 0.779 (+0.0\%) \\ \hline
\end{tabular}
\end{table}

\subsection{Investigation on different building heights and spatial resolutions}\label{sec:disc:height}
\begin{figure}[!ht]
  \centering
  \includegraphics[width=\linewidth]{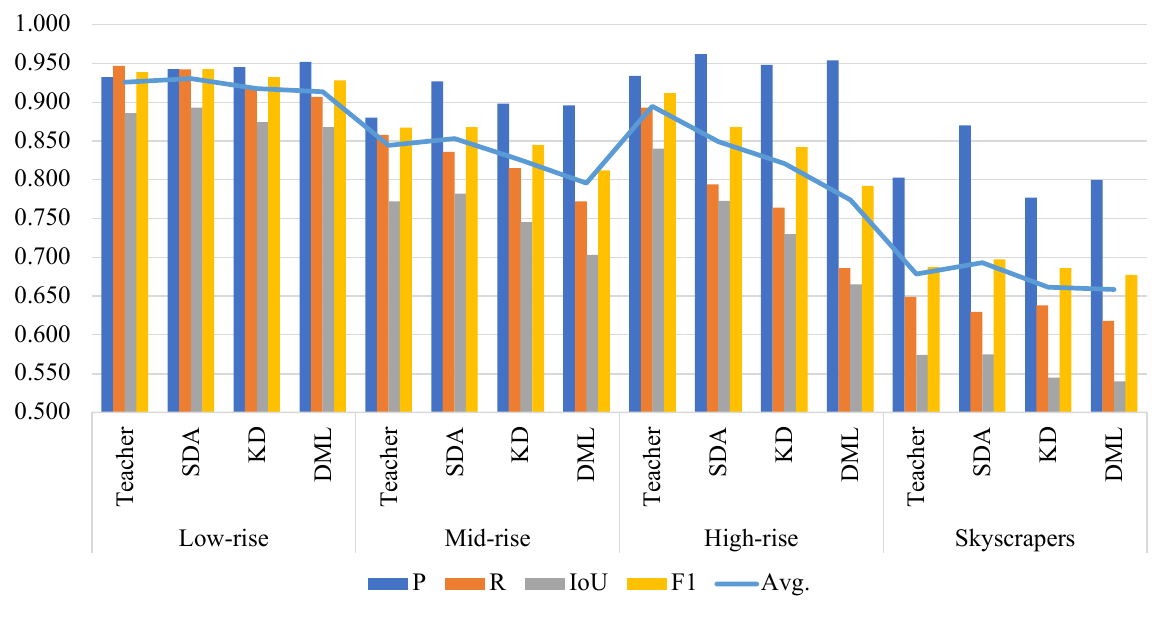}
  \caption{The segmentation results of the Teacher, SDA, KD, and DML using the U-VGG19 network on four types of buildings: low-rise, mid-rise, high-rise, and skyscrapers. The highest scores for each building type are highlighted in bold. ``Avg.'' refers to the average of the four evaluation scores.}
      \label{fig:buildingtype-results}
      \end{figure}

The effects of building heights were studied by separating the image scenes of Ev into four categories of building types: low-rise, mid-rise, high-rise, and skyscrapers. The chart in Figure \ref{fig:buildingtype-results} shows the performance of SDA as compared to the Teacher, KD, and DML. SDA outperformed the Teacher, KD, and DML with the highest scores for low-rise (0.943 F1), mid-rise (0.868 F1), and skyscrapers (0.697 F1). Both Student and Teacher networks yielded F1 and average scores above 90\% for low-rise buildings. The performance degraded with the increased height of the buildings.

\begin{figure*}[!ht]
  \centering
  \includegraphics[width=\linewidth]{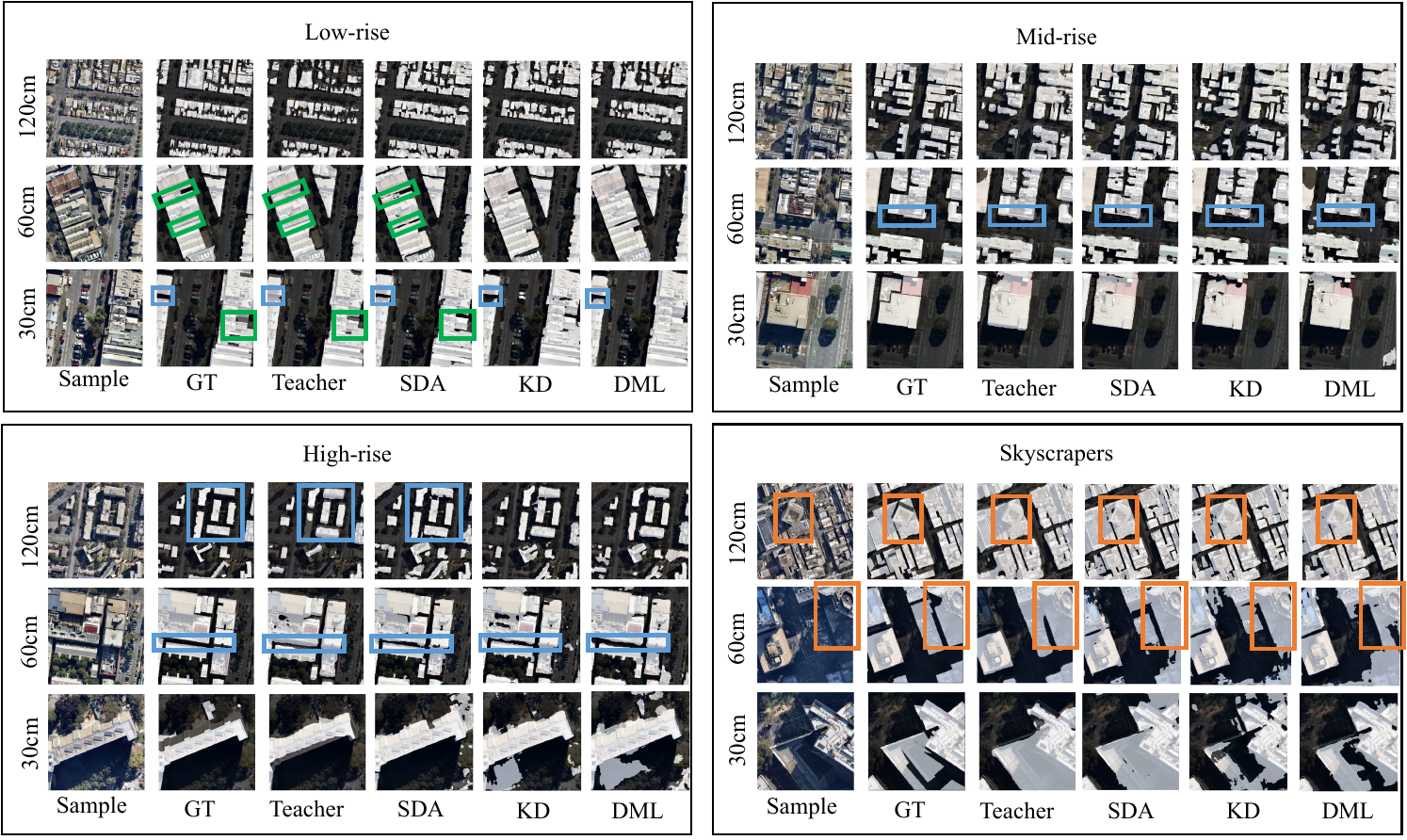}
  \caption{Result samples on four types of buildings and samples of different spatial resolutions. Green boxes highlight how the Student networks can separate the instances of separated buildings. The blue boxes show that the methods are trying not to detect the facades of the buildings. Orange boxes are used to highlight the skyscraper buildings in the provided samples. Each 30, 60, and 120 cm sample shows skyscrapers of 177m, 251m, and 319m (the tallest building in Melbourne).}
      \label{fig:mainresult}
      \end{figure*}

The study was further separated with the spatial resolution resolution of the images. Figure \ref{fig:mainresult} shows sample output from each method with images of three spatial resolutions for the four building types. 
For the low-rise buildings, SDA increased the generalisation of the buildings in all three resolutions and achieved the highest F1 of 0.943. The green boxes highlight the network's ability in instance segmentation of buildings. The blue boxes in the 30 cm images show that SDA could separate the roofs from the facades. KD and DML over-segmented the images. 
For mid-rise, the performance was lower than low-rise with SDA showing some improvements. The blue boxes in the 60 cm image sample show SDA's ability to segment out the facades and the gaps between the buildings.
For the high-rise, the Teacher performed with the highest evaluation scores despite the noisy training data. However, it was seen that SDA could separate roofs from the facades with better segmentation.
For skyscrapers (outlined with orange boxes), the segmentation was poor. SDA produced the highest scores but with increased noise. The 30 cm sample in Figure \ref{fig:mainresult} shows the segmentation of the tallest skyscraper in Melbourne (319m high).

The spatial resolution of the images affected the segmentation of the four building types differently. From Figure \ref{fig:mainresult}, the Students performed poorly on the highest resolution (30 cm) images. Low-rise buildings were too small to segment accurately, and high-rise buildings and skyscrapers were too large to fit into the image tiles. The best visual results were achieved at the resolution of 60 cm, except for skyscrapers, for which the lowest resolution of 120 cm provided the best segmentation.

\subsection{Applications and challenges}\label{challenges}
The urban building extraction methods developed in this paper are fundamental for creating inventories supporting various applications. This includes urban base maps, urban planning, sprawl management, disaster management, environmental monitoring (such as urban heat islands and green space management), real estate market analysis, navigation, and climate change research. These building inventories directly enhance societal aspects of urban living, disaster preparedness, economic growth, environmental conservation, and social equity.

\begin{figure}[!ht]
  \centering
  \includegraphics[width=\linewidth]{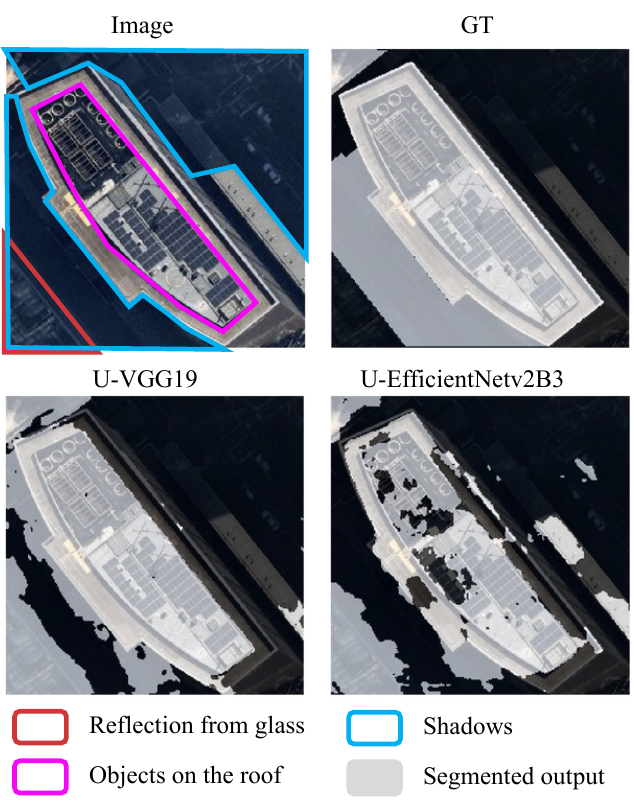}
  \caption{Sample data and segmentation outputs of U-VGG19 and U-EfficientNetv2B3 after SDA on an image scene with a skyscraper. The results are affected due to several challenges posed in urban scenes with tall buildings.}
      \label{fig:limitations}
      \end{figure}

However, extracting urban buildings like high-rises and skyscrapers presents significant challenges. These include shadows cast by tall buildings, reflections from glass facades, and rooftop objects. Tackling the misalignment between building labels and images adds to these challenges. Figure \ref{fig:limitations} demonstrates the segmentation outputs from two top-performing Student networks from our study on an image tile containing a skyscraper. The models struggled with accurate segmentation due to these challenges. In Figure \ref{fig:mainresult}, SDA improved segmentation by reducing the over-segmentation of facades and gaps between nearby buildings. Nevertheless, accurately segmenting individual building instances remains difficult.

Beyond segmentation, deploying knowledge distillation in operational systems poses challenges due to storing multiple networks in memory. In contrast, adapting a pre-trained network with SDA proved easier to deploy. Given the importance of building inventories for numerous applications and societal benefits, it is essential to improve the processes for extracting and updating them promptly.

\section{Conclusion}
This paper proposed supervised domain adaptation to train noise-tolerant models for the problem of misaligned building labels due to off-nadir aerial imagery. SDA was studied with three new datasets we developed and several EDNs that we configured using lightweight CNNs from computer vision. The developed datasets consisted of image-label pairs with ``large-noisy'' and ``small-clean'' labels to facilitate the training of noise-tolerant models. 

The investigation benchmarked 43 lightweight CNNs, 7 EDNs, 5 optimisers and 9 loss functions. The best hyperparameters to train an EDN were identified as RMSProp optimiser, Dice loss, and 0.0001 learning rate, which yielded up to 0.962 F1 and 0.039 loss. The best Students included U-EfficientNetv2B3 (with 0.965 F1) and U-EfficientNet-lite0 with the best trade-off between the scores (0.962 F1), network parameters, and training time. The best Teacher was U-VGG19 (0.967 F1) which opted for a similar trade-off. Thus identified Teacher and Students were used to study SDA for the misalignment problem. Compared to the existing methods like KD and DML, SDA was the most effective method to address the problem for all building types with the highest F1 in low-rise (0.943), mid-rise (0.868), high-rise (0.912), and skyscrapers (0.697). Building height and spatial resolution of aerial images strongly affected the segmentation performance from SDA and other methods. Among multiple spatial resolutions studied (30, 60, and 120 cm), 60 cm images provided the best trade-off between local and global context information for low-rise, mid-rise, and high-rise buildings.

The experiments produced key insights into training noise-tolerant neural networks to address misalignment in building labels and images caused by off-nadir source imagery. This problem, encountered by providers like Microsoft, Google, ESRI, and Geoscape, will benefit from these findings, enhancing the accuracy of urban building inventories.

\section*{Acknowledgement}
The first author (B.N.) is supported by the University of Melbourne for his PhD research and is awarded the Melbourne Research Scholarship. The authors would like to thank Nearmap for providing the API service to collect the image data for the experiments.

\bibliographystyle{apalike}
\bibliography{references}

\end{document}